# A Bag of Visual Words Model for Medical Image Retrieval

**Sowmya Kamath S** [1]      and      **Karthik K** [2]

1 Assistant Professor, Department of Information Technology, National Institute of Technology Karnataka, Surathkal, Mangalore 575025, India, e-mail: *sowmyakamath@nitk.edu.in*
2 Research Scholar, Department of Information Technology, National Institute of Technology Karnataka, Surathkal, Mangalore 575025, India. email: *2karthik.bhat@gmail.com*

**ABSTRACT:** Medical Image Retrieval is a challenging field in Visual information retrieval, due to the multi-dimensional and multi-modal context of the underlying content. Traditional models often fail to take the intrinsic characteristics of data into consideration, and have thus achieved limited accuracy when applied to medical images. The Bag of Visual Words (BoVW) is a technique that can be used to effectively represent intrinsic image features in vector space, so that applications like image classification and similar-image search can be optimized. In this paper, we present a MedIR approach based on the BoVW model for content-based medical image retrieval. As medical images as multi-dimensional, they exhibit underlying cluster and manifold information which enhances semantic relevance and allows for label uniformity. Hence, the BoVW features extracted for each image are used to train a supervised machine learning classifier based on positive and negative training images, for extending content based image retrieval. During experimental validation, the proposed model performed very well, achieving a Mean Average Precision of 88.89% during top-3 image retrieval experiments.

## INTRODUCTION

Medical Image retrieval is a very challenging and active research with applications in similarity based automated diagnosis, decision support and medical data management (Niblack et al, 1993). Content based image retrieval (CBIR) and keyword based querying are the most popular methods in Medical Image Retrieval (MedIR). In text/keyword based querying, traditional database systems and text based annotations stored along with each image are used for facilitating retrieval. In CBIR, the objective is to find more images similar to a given query image, based on their actual visual content (and not by any metadata associated with it), and rank them as per their similarity with the query image. CBIR utilizes image-level features like colour, shape and texture for identifying relevant images for retrieval, with reference to the given query image. Even though keyword based matching and retrievals are fast, it is only viable when medical images are sufficiently annotated. Modern medical diagnostic tools contribute varied types of medical images, like X-rays and CAT scans, which are huge in volume and continuously generated, due to which effective textual annotations are few. When available, these annotations are often ambiguous due to the unstructured nature of natural language or incomplete, thus adversely affecting retrieval results. The Picture Archival and Communication (PACS) System (Lehmann et al, 2003; 2004) is a significant effort to overcome these challenges, with a focus on effectively storing, retrieving and transmitting medical images. However, a major limitation is that, it uses techniques that rely only on keywords and associated text annotations stored along with the image for retrieval. This later led to the development of and subsequent popularity of CBIR systems specifically for medical image management.

CBIR systems focus on capturing the latent features of an image dataset without relying on any external information (e.g. text metadata associated with images). Most CBIR systems leverage features like colour and texture for producing relevancy rankings. However, a significant hurdle is, as most medical images are greyscale, colour cannot be used as a prominent feature. But, image texture and shape are crucial in the case of medical images, which need to be effectively captured.

Content based image retrieval was popularized by the ASSERT system (Shyu et al, 1999) can perform automated indexing of high-resolution computed tomography (CT) scans of lungs in medical imaging. By utilizing the grayscale and texture attributes from the image co-occurrence matrix for characterization of the dataset, the system allowed extraction of pathology-bearing regions in lung scans. The web based IRMA (Information Retrieval for Medical Applications) (Keysers et al, 2003) system accepts a X-ray image as query, for discovery of similar images from the database. Here, the image retrieval process is broken down to seven steps where every step represents a hierarchy of image abstraction that signifies a high level understanding of the image content. To compute the closest image for a given query, the local textures and similarity measures are incorporated. Another technique, Flexible Image Retrieval Engine (FIRE) (Deselaers, 2004) includes nonmedical datasets like photographic databases and utilizes feature engineering to dynamically adapt weights assigned to every feature during retrieval. Numerous CBIR ranking policies based on IR models like BIM (Binary Independence Model) [6], BM25 (Best Match Okapi 25) [7], Vector Space Model (VSM) [8] etc have been developed over the years.

In this paper, we propose the application of the Bag of Visual Words (BoVW) weighting model for supporting similar image search over a large-scale image database of X-Ray images, i.e., Content-based Medical Image Retrieval (CBMIR). The visual vocabulary is constructed using the image features extracted using the SURF feature extraction algorithm. K-means algorithm was applied to reduce the number of irrelevant features and prune the visual vocabulary. Machine learning classifiers were trained on the feature representations of the images for improved precision during retrieval w.r.t to a user's query. The rest of the paper is organized as follows: Section 2 discusses the proposed methodology in detail. The experimental setup and observations are presented in Section 3, followed by conclusion and possible directions for future work.

## PROPOSED METHODOLOGY

The objective of our work is to enable similar image retrieval when a sample query image is provided by medical personnel like doctors and radiologists. Fig. 1 shows the proposed methodology in detail. For the experiments, a standard dataset popularly called the ImageCLEF 2009 challenge dataset (Tommasi et al, 2009) containing more than 12,000 X-ray images of the abdomen, chest, hand, shoulder, face etc was used. We used a total of 47 different classes for our experiments. Each image category contains equal number of images, which were first separated into training and testing, as per the Using 70:30 ratio.

Creation of Visual Vocabulary is a major process which helps in effective representation of the monochromatic X-ray images in a concept feature space. As images do not include discrete words, a 'vocabulary' of visual words is constructed by extracting image level features from each image across the various categories. Firstly, 70% of images from each set was considered as training data and the remainder 30% as test data. The SURF algorithm (Speeded-up Robust Features) (Bay et al, 2008) was applied to both training and test sets to find visually interesting points and encode information about the area around these points as its representative feature vector. We used SURF as it provides greater scale and rotation invariance when compared to other feature extraction, and hence may be more suitable for X-ray images, which are often prone to misalignment. Each extracted feature is a discernible and significant point/group of points in an image, like, its corner points, edges, blobs, contours etc.

Next, the visual vocabulary is constructed by minimizing the number of features through quantization. To achieve this K-means clustering was applied to the bag of features, so that clusters of similar image data points are obtained. In the first iteration, random cluster centroids are chosen, and the algorithm iterates over each input feature and assigns them to the similar centroid based on the standardized Euclidean distance measure. During the next iteration, each centroid is updated so that it is more tightly aligned with the newly formed cluster and the feature assignment is again performed. This process is continued until no changes in centroid position are observed from the previous iteration, which is considered to be the best possible clustering.

After this process, only 80% of the strongest features of each image in the bag of visual words are retained, which forms the training data. Each bag of features (BoF) object provides an encoded scheme for counting the visual word occurrences in an image using which a histogram that represents the reduced feature set of an image is generated (shown in Fig.2). This reduced feature set is used as a training set for training a machine learning classifier for medical image classification. After this process, each image in the training and test datasets is represented as an encoded feature vector. Encoded training images from each category are then fed into a classifier for training.

We used the Image Category classifier available in Matlab 2014Ra for image classification, which is fed the number of categories and the category labels for the input medical images. The function trains a support vector machine (SVM) multiclass classifier using the input bag, a BoF object. The trained classifier is applied to categorize the new images in the test set, for predicting the label index value. Using the provided labels and the predicted label, standard information retrieval (IR) metrics were used to analyze the performance of the proposed MedIR technique, which is presented in the next section.

**EXPERIMENTAL RESULTS**

The proposed MedIR model was implemented using the Matlab 2014Ra setup. A highend workstation with Intel Xeon Octacore processor of speed 3.31 GHz and 16 GB of RAM was used for the implementation. For experimental validation, the labelled data available for every image in the ImageCLEF 2009 dataset was used. Approximately 12,000 labelled X-ray images of multiple organs such as spine, palm bones, knees, ankles, etc, were used for experiments. Fig.3 shows a representative sample of the ImageCLEF dataset.

For evaluation of the retrieval results, we considered a metric called Precision@$k$. It is defined as the amount of relevant results in the first $k$ search results for the given query. In the considered setting, precision@$k$ is the number of images of the same category as that of the query image in the first $k$ retrieval results. Three cases, $k=3$, 5 and 10 were considered for experimental evaluation of retrieval results for a set of sample queries (images). Mean Average Precision (MAP) is another metric which can be used to evaluate overall system performance. MAP@$k$ is the mean of the precision at $k$ values obtained for various query images, and is given by Eq.1.

$$MAP_k = 1/Q \sum_{q=1}^{q=Q} P_k(q) \qquad (1)$$

Where, Q is the set of $q$ queries used for the retrieval and $P_k(q)$ is the precision@$k$ value observed for that particular query $q$.

As can be seen from Table 2, the value of MAP@k=3 shows that the top-3 similar image retrieval was very effective with the highest MAP value of 88.89%. As the value of k is varied, it can be seen that retrieval performance degrades, the lowest being 70% when k=10. Considering the fact that most medical image retrieval applications require best performance with low recall, it can be concluded that the proposed system is well-suited for real-world applications.

## CONCLUSIONS

The following conclusions are deduced from this study:
- A BoVW based MedIR model was proposed for large-scale similar medical image retrieval.
- The ImageCLEF 2009 dataset was considered for experimental evaluation with standard metrics like Precision@$k$ and Mean Average Precision (MAP).
- The proposed MedIR approach achieved a MAP@$k$=3 performance of 88.89%, indicating that it is very suitable for real-world similar medical image management and retrieval applications.

As future work, the applicability of vocabulary pruning techniques like Probabilistic Latent Semantic Analysis (Hoffman et al, 1999) and Principal Component Analysis (Joliffe, 1986) will be explored, for optimal representation of medical images, for improved performance at higher values of $k$.

## ACKNOWLEDGEMENTS

We gratefully acknowledge the financial support and the facilities at the Department of Information Technology, NITK Surathkal, funded by the DST-SERB Early Career Research Grant to the first author (ECR/2017/001056).

## REFERENCES

[1] Ashnil Kumar, Jinman Kim, Weidong Cai, Michael Fulham, Dagan Feng, ―Content-Based Medical Image Retrieval: A Survey of Applications to Multidimensional and Multimodality Data‖, J Digit Imaging (2013) 26: 1025

[2] Bay, H., Ess, A., Tuytelaars, T., & Van Gool, L. (2008). Speeded-up robust features (SURF). Computer vision and image understanding, 110(3), 346-359.

[3] Deselaers, T., Keysers, D & Ney, H. (2004, September). FIRE-flexible image retrieval engine: ImageCLEF 2004 evaluation. In CLEF (pp. 688-698)

[4] Hofmann, Thomas. "Probabilistic latent semantic analysis." Proceedings of the Fifteenth conference on Uncertainty in artificial intelligence. Morgan Kaufmann Publishers Inc., 1999.

[5] Jolliffe, Ian T. "Principal component analysis and factor analysis." Principal component analysis. Springer, New York, NY, 1986. 115-128.

[6] Keysers, D., Ney, H., Wein, B. B., Lehmann, T. M. (2003). Statistical framework for model-based image retrieval in medical applications. Journal of Electronic Imaging, 12(1), 59-68.

[7] Lehmann, T. M., Gold, M. O., Thies, C., Fischer, B., Spitzer, K., Keysers, D & Wein, B. B. (2004). Content-based image retrieval in medical applications. Methods of information in medicine, 43(4), 354-361.

[8] Lehmann, T. M., Guld, M. O., Thies, C., Fischer, B., Keysers, D., Kohnen, M., ... & Wein, B. B. (2003, May). Content-based image retrieval in medical applications for picture archiving and communication systems. In Medical Imaging 2003: PACS and Integrated Medical Information Systems: Design and Evaluation (Vol. 5033, pp. 109-118). International Society for Optics and Photonics.

[9] Lowe, D. G. (2004). Distinctive image features from scale-invariant keypoints. International journal of computer vision, 60(2), 91-110.

[10] Niblack, C. W., Barber, R., Equitz, W., Flickner, M. D., Glasman, E. H., Petkovic, D., ... & Taubin, G. (1993, April). QBIC project: querying images by content, using color, texture, and shape. In Storage and retrieval for image and video databases (Vol. 1908, pp. 173-188). International Society for Optics and Photonics.

[11] Shyu, C. R., Brodley, C. E., Kak, A. C., Kosaka, A., Aisen, A. M., & Broderick, L. S. (1999). ASSERT: A physician-in-the-loop content-based retrieval system for HRCT image databases. Computer Vision and Image Understanding, 75(1-2), 111-132.

[12] Tommasi, T., Caputo, B., Welter, P., Güld, M. O., & Deserno, T. M. (2009, September). Overview of the CLEF 2009 medical image annotation track. In Workshop of the Cross-Language Evaluation Forum for European Languages (pp. 85-93). Springer, Berlin, Heidelberg.

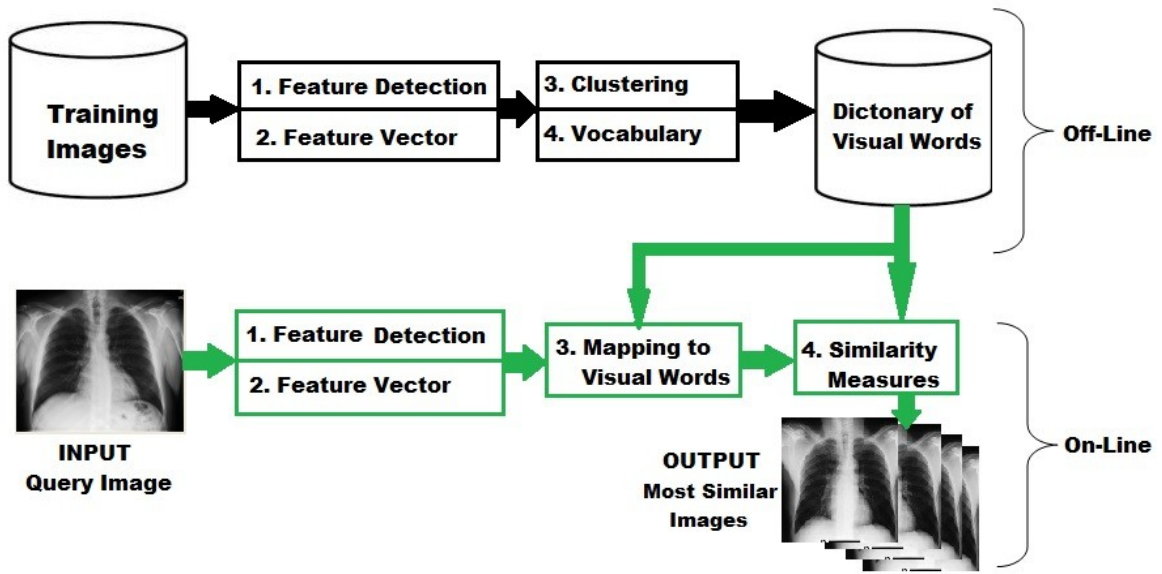

**Fig.1 Proposed MedIR methodology**

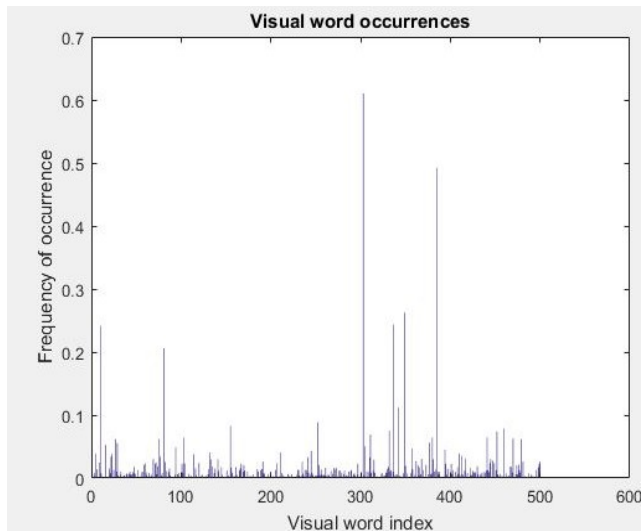

**Fig 2. Histogram of visual word occurrences**

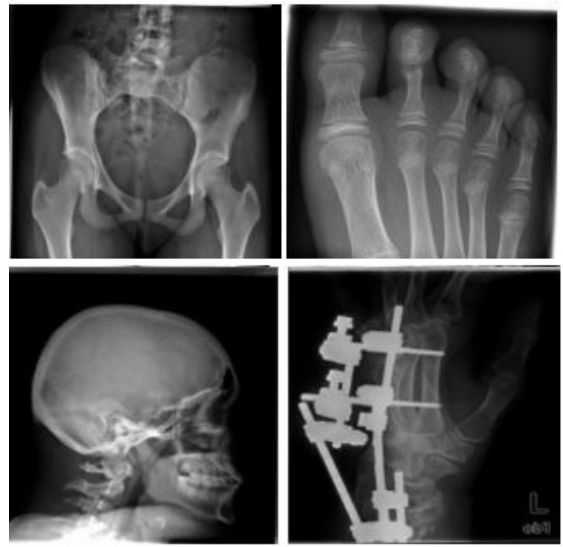

**Fig.3 ImageCLEF 2009 sample data**

**Table 2. Evaluation of retrieval with Precision@*k* for *k=3, 5, 10***

| Query (images) | k value | No. of relevant images at k | Precision@k |
|---|---|---|---|
| Sample 1 (Chest class) | k=3 | 2 | 66.67% |
| | k=5 | 3 | 60% |
| | k=10 | 8 | 80% |
| Sample 2 (Lumbar Spine class) | k=3 | 3 | 100% |
| | k=5 | 4 | 80% |
| | k=10 | 5 | 50% |
| Sample 3 (Hand class) | k=3 | 3 | 100% |
| | k=5 | 4 | 80% |
| | k=10 | 8 | 80% |

**Table 3. Evaluation of retrieval with MAP@*k* for *k=3, 5, 10***

| k value | MAP@k |
|---|---|
| k=3 | 88.89% |
| k=5 | 73.33% |
| k=10 | 70% |

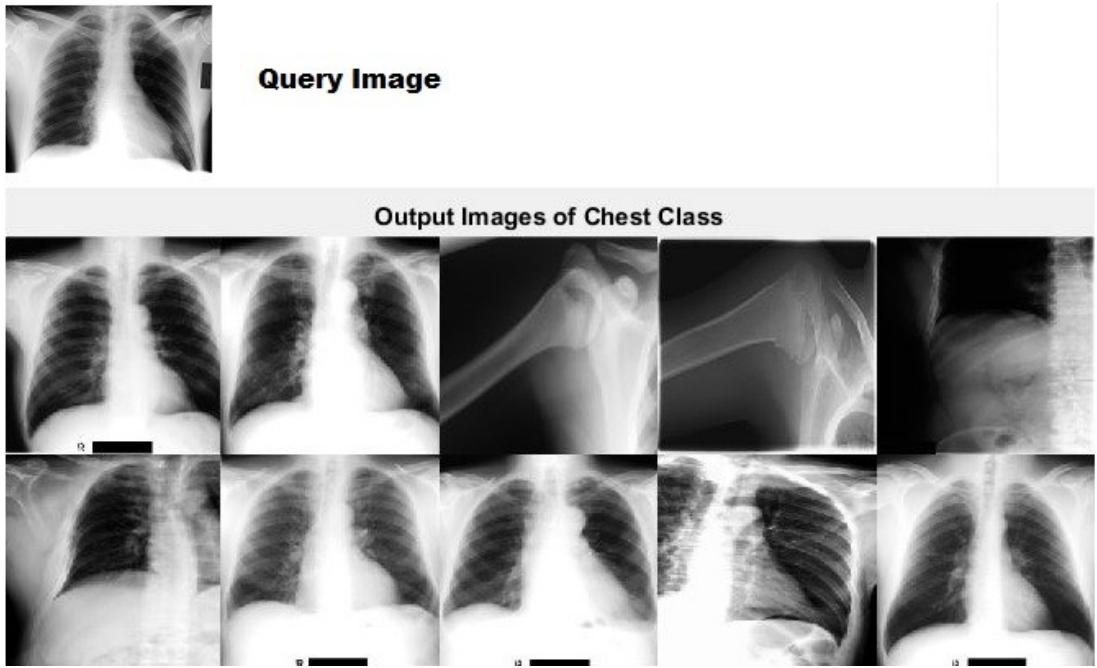

**Fig.3 Observed retrieval results for Chest X-ray class**

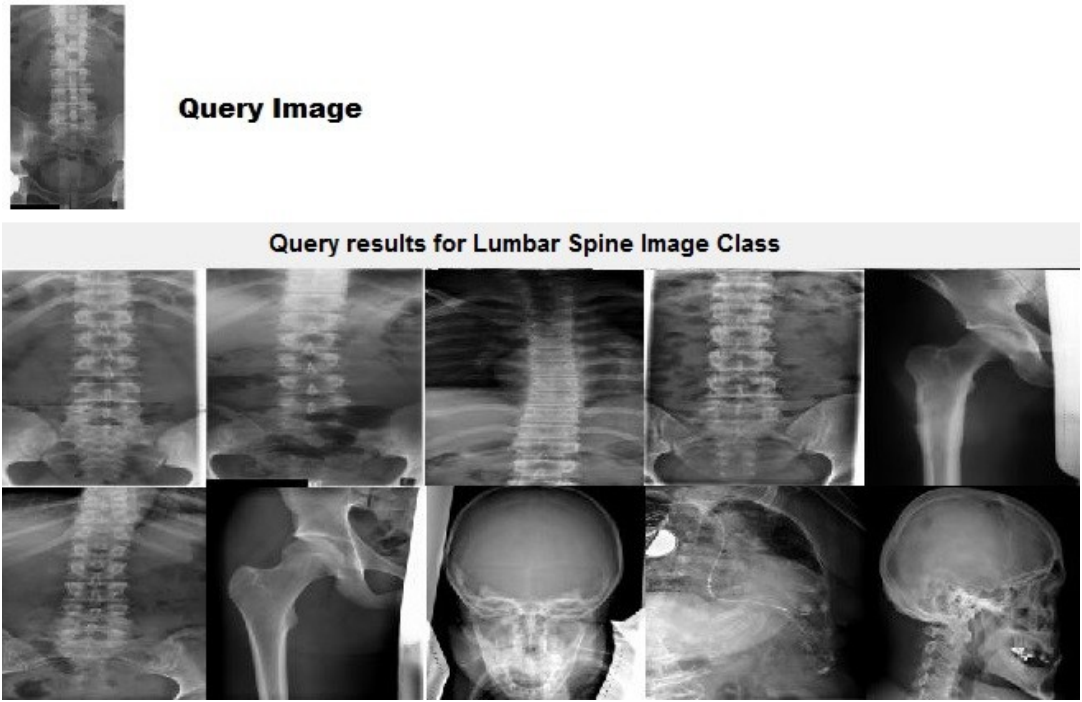

**Fig.4 Observed retrieval results for Lumbar spine X-ray class**